\documentclass[10pt,twocolumn,letterpaper]{article}

\usepackage[pagenumbers]{cvpr}              

\usepackage{graphicx}
\usepackage{amsmath,amssymb,amsfonts}
\usepackage{booktabs}
\usepackage{multirow}
\usepackage{tabularx}
\usepackage{array}
\usepackage[table]{xcolor}
\usepackage{colortbl}
\usepackage{xspace}
\usepackage{tikz}
\usetikzlibrary{positioning, fit, arrows.meta, calc, backgrounds}
\usepackage{algorithm}
\usepackage{algorithmic}
\newcommand{\method}{PerBite\xspace}

\definecolor{TableHeader}{HTML}{EEF3F8}
\definecolor{TableRule}{HTML}{2F5E85}
\definecolor{TableStripe}{HTML}{F8FAFC}
\definecolor{inputbg}{HTML}{EBF3FF}
\definecolor{inputtext}{HTML}{1A4780}
\definecolor{corebg}{HTML}{E6F9EC}
\definecolor{coretext}{HTML}{13532B}
\definecolor{calibbg}{HTML}{FFF3E6}
\definecolor{calibtext}{HTML}{663C00}
\definecolor{auxbg}{HTML}{F2F2F7}
\definecolor{auxtext}{HTML}{48484A}
\newcommand{\thead}[1]{\textbf{#1}}
\newcommand{\metricdown}{\,$\downarrow$}
\newcommand{\modernrule}{\arrayrulecolor{TableRule}\midrule\arrayrulecolor{black}}
\renewcommand{\arraystretch}{1.12}

\definecolor{cvprblue}{rgb}{0.21,0.49,0.74}
\usepackage[pagebackref,breaklinks,colorlinks,allcolors=cvprblue]{hyperref}

\def\paperID{}
\def\confName{CVPR}
\def\confYear{2026}

\title{PerBite: A Curated Diagnostic Workflow for Bite-Aware Food Volume Estimation}

\author{Ahmad AlMughrabi$^{*,1}$\\
{\tt\small ahmad.almughrabi@ub.edu}
\and 
Farid Al-Areqi$^{*,1}$\\
{\tt\small farid.al-areqi@ub.edu}
\and
David Fernández Gómez$^{*,1}$\\
{\tt\small dfernago24@alumnes.ub.edu}
\and
Umair Haroon$^{1}$ \\
{\tt\small umairharoon@ub.edu}
\and
Marc Bolaños$^{4}$ \\
{\tt\small marc.bolanos@logmeal.com}
\and
Ricardo Marques$^{\infty,2}$\\
{\tt\small ricardo.marques@upf.edu }
\and 
Petia Radeva$^{\infty,1,3,4}$\\
{\tt\small petia.ivanova@ub.edu}
}

\begin{document}
\twocolumn[{
\renewcommand{\thefootnote}{}%
\maketitle
\vspace{-3.0em}
\begin{center}
\includegraphics[trim={0 5cm 0 0},clip,width=0.90\textwidth]{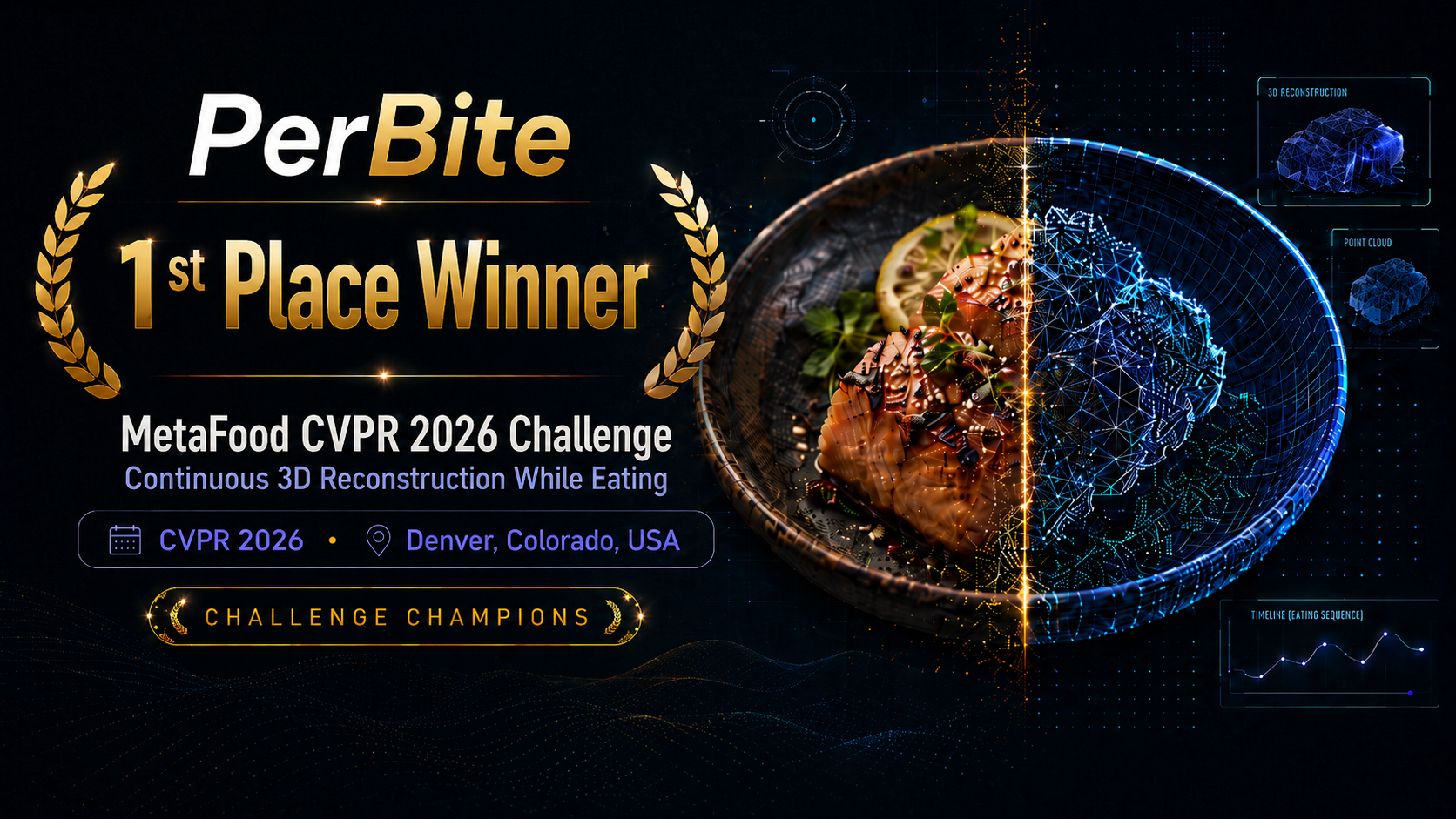}
\end{center}
\vspace{0.3em}
}]

\def\thefootnote{$^*$}\footnotetext{These authors contributed equally to this work.}
\def\thefootnote{$^\infty$}\footnotetext{Equal supervision.}
\def\thefootnote{$^1$}\footnotetext{Universitat de Barcelona, Spain.}
\def\thefootnote{$^2$}\footnotetext{Grup de Tecnologies Interactives, Universitat Pompeu Fabra, Spain.}
\def\thefootnote{$^3$}\footnotetext{Institut de Neurociències, Barcelona.}
\def\thefootnote{$^4$}\footnotetext{LogMeal, AIGecko Technologies S.L.}

\setcounter{footnote}{0}
\renewcommand{\thefootnote}{\arabic{footnote}}

\begin{abstract}

Can a visually plausible food mesh be trusted to estimate the volume of consumed food? \method investigates this question using selected paired before- and after-consumption states from the MetaFood CVPR 2026 Continuous 3D Reconstruction While Eating Challenge. The submitted workflow follows a curated reconstruction protocol: SAM~3 segments the food and plate regions; Hunyuan3D/SAM~3D generates a dimensionless food mesh; the plate diameter provides the metric scale; the plate geometry is removed in Blender; and the remaining mesh is hole-filled, made watertight, and integrated to estimate volume. MoGe-2 is used only as an auxiliary cue for initial dish-diameter estimation when direct plate measurement is uncertain; it is not the primary scale source for the reported challenge result. \method ranks first, with an average Chamfer distance of 8.31 across 34 meshes using rigid ICP without scale correction. On 17 before- and after-pairs, it achieves 33.87\% state-level volume MAPE and zero monotonicity violations, while consumed-volume MAPE remains 53.74\%. The results show that surface reconstruction, metric scale, controlled mesh cleanup, watertight volume integration, and physical depletion consistency should
be evaluated separately for dietary assessment. Source code and evaluation scripts will be available at \href{https://github.com/GCVCG/PerBite-CVPR-MetaFood-2026}{github.com/GCVCG/PerBite-CVPR-MetaFood-2026}.
\end{abstract}

\section{Introduction}
\label{sec:intro}

Image-based dietary assessment has historically focused on estimating food type and portion size from one or a small number of images~\cite{almughrabi2024voleta,haroon2026vole,almughrabi2025voltex}. This setting is convenient, but it often depends on users deliberately capturing suitable images and provides only sparse evidence of the eating process. As a result, it cannot fully capture how food changes during consumption, how much is ultimately eaten, or the meal's temporal progression. Continuous egocentric video provides a more natural signal because the camera observes the meal throughout consumption, enabling more passive estimation of intake dynamics. However, it also introduces severe motion blur, viewpoint changes, occlusions by hands or utensils, and partial visibility. The core challenge is therefore not simply to reconstruct a visually plausible 3D model, but to recover metrically meaningful food volumes that remain consistent across consumption states.

Bite-aware intake estimation is particularly difficult because the quantity of interest is not only the volume of food visible at a single moment, but the change in food volume across multiple consumption states. Small errors in scale, boundary localization, support removal, or mesh closure can propagate into large errors in consumed volume when intake is computed as the difference between two state estimates. This makes continuous dietary monitoring a paired-state reasoning problem, where remaining-volume estimates must be metrically calibrated and physically consistent over time.

We present \method, a curated and bite-aware diagnostic workflow for \emph{metric-calibrated food volume estimation while eating}. Given selected before- and after-consumption food states, \method estimates remaining volumes $\{V_t\}_{t=0}^{T}$ and consumed increments $\Delta V_t=V_{t-1}-V_t$. At a high level, the workflow combines promptable food-state localization, single-image 3D reconstruction, metric scale calibration, non-food geometry removal, mesh repair, and volume integration. This paper focuses on the reconstruction, scaling, cleanup, and volume-differencing problem; it does not claim automatic bite detection or automatic before/after frame selection.

PerBite was evaluated as part of the MetaFood CVPR 2026 Continuous 3D Reconstruction While Eating Challenge\footnote{\url{https://www.kaggle.com/competitions/mtf-challenge-1-continuous-3-d-reconstruction-while-eating}}, where it ranked first under the official challenge protocol. This external evaluation is important because it tests the method under a fixed benchmark rather than a self-defined split. In this paper, we go beyond the challenge score by analyzing surface reconstruction, state-level volume accuracy, consumed-volume error, and monotonic depletion consistency.

We organize the paper around three diagnostic questions: (i) whether a compact, promptable reconstruction path can be competitive when supported by controlled mesh cleanup, (ii) how metric evidence makes a generated unitless mesh volume meaningful, and (iii) why consumed volume is harder than remaining volume. These questions motivate the design of the method and the metrics reported below.

The central design principle is to treat bite-aware food-volume estimation as \emph{paired-state metric reasoning}, rather than as independent per-frame reconstruction. Selected state images define the before/after inputs, metric calibration determines how reconstructed meshes are converted into physical units, geometry cleanup determines which surface is finally integrated, and monotonic depletion provides a simple check on whether the predicted trajectory is physically plausible. This framing is aligned with recent work on egocentric dietary tracking and diagnostic volume benchmarking: temporal signals are essential for bite-level intake, and volume accuracy depends on interactions among support, depth, scale, mesh cleanup, and integration rather than on visual plausibility alone~\cite{almughrabi2025voltex,haroon2026vole,onevol2026}.

We frame \method as both a reconstruction method and a diagnostic protocol. The proposed evaluation decouples surface reconstruction quality from volume correctness: a mesh can achieve an acceptable Chamfer distance while still producing poor volume if the global scale, object boundary, or bitten region is incorrect. This distinction is central to continuous dietary monitoring, where the consumed volume is the difference between two noisy state estimates and is therefore more fragile than the remaining volume in a single state. Our main contributions are:

\begin{itemize}
    \item \textbf{A bite-aware diagnostic workflow.} We introduce \method for paired food-state reconstruction and consumed-volume estimation, validated in the MetaFood CVPR 2026 challenge setting.
    \item \textbf{A curated promptable reconstruction protocol.} \method combines promptable food-state localization, generated 3D mesh reconstruction, non-food geometry removal, and controlled mesh repair before volume computation.
    \item \textbf{Metric scaling for generated food meshes.} We recover physical food volume by applying a uniform metric scale before measuring the repaired food mesh volume.
    \item \textbf{Bite-level diagnostic evaluation.} We evaluate before/after food states with no-scale-correction Chamfer distance, state-level volume error, consumed-volume error, and monotonic depletion consistency.
\end{itemize}

\section{Related Work}
\label{sec:related}

This section positions \method relative to food-volume estimation, egocentric intake monitoring, promptable reconstruction, and metric-scale reasoning. The key distinction is that \method evaluates pairs of consumption states and separates unitless shape generation from metric volume calibration.

\subsection{Image-based Food Volume Estimation}
\label{sec:related_food_volume}

Food-volume estimation has been studied using reference objects, stereo or multi-view reconstruction, RGB-D sensing, geometric primitives, and learned depth or segmentation models. Early and related systems demonstrate that physically meaningful portion estimation depends not only on recognizing the food category, but also on recovering metric geometry and scale. For example, wearable-camera-based approaches have explored 3D reconstruction for estimating food volume in dietary intake monitoring~\cite{Gao2018FoodVE}, while other image-based methods use controlled capture, depth cues, or geometric assumptions to make, portion estimates metrically interpretable~\cite{almughrabi2024voleta,almughrabi2025voletapp,haroon2026vole,haroon2025volepp,almughrabi2025voltex}. Recent monocular methods reduce the capture burden by leveraging learned depth, segmentation, video segmentation, or image-to-3D priors~\cite{almughrabi2025foodmem,almughrabi2026benchseg}, yet monocular scale ambiguity and object-boundary leakage remain major failure modes. Our work follows this metric tradition but shifts the problem from static pre-meal estimation to paired pre- and post-consumption food states.


\subsection{Challenge Datasets and Diagnostic Protocols}
\label{sec:related_challenges}

Recent challenge datasets have helped standardize evaluation for physically informed food reconstruction. The MetaFood CVPR 2024 challenge evaluated volume-accurate 3D food reconstruction from 2D images using a visible reference pattern for scale~\cite{he2024metafood}. This line of work is important because it exposes the gap between visually plausible reconstruction and physically accurate estimates of food volume. The current MetaFood continuous eating setting extends this difficulty to before- and after-consumption states, where errors in scale, support removal, or bitten-region geometry can directly affect estimates of consumed volume. We adopt a diagnostic view: volume estimation should be evaluated not only by surface similarity, but also by metric scale, object boundary quality, mesh integration, and before/after consistency.


\subsection{Egocentric Dietary Monitoring}
\label{sec:related_ego}

Egocentric and wearable cameras provide a natural sensing modality for dietary monitoring because they can observe the meal from the user's perspective with less active intervention than manually captured food images. Prior work using chest-worn cameras has demonstrated the relevance of passive image capture for food portion size estimation and dietary assessment~\cite{jia2014accuracy}. Wearable-camera food-volume estimation further demonstrates the potential of egocentric sensing for quantifying intake through reconstructed food geometry~\cite{Gao2018FoodVE}. However, egocentric dietary assessment remains challenging because food may be partially visible, occluded by hands or utensils, and observed under changing viewpoints. Our method differs from these works by explicitly reconstructing and metrically scaling paired food states, while adopting the same core motivation: intake estimation benefits from observing the eating episode rather than relying solely on isolated single-image predictions.


\subsection{Promptable Segmentation and 3D Reconstruction}
\label{sec:related_promptable}
SAM~3 extends promptable segmentation to concept prompts, including text prompts that can detect, segment, and track all matching instances in images and videos~\cite{carion2025sam3}. This capability allows \method to use a single generic prompt, ``food'', for localization. SAM~3D reconstructs 3D objects from single images~\cite{sam3d2025}, making it a natural unitless shape prior for paired food-state reconstruction. We emphasize, however, that generated meshes are not automatically metric; metric scale must be recovered from observed reference or geometric cues.

\subsection{Metric Scale Recovery}
\label{sec:related_metric}
Metric scale is a central difficulty in monocular food reconstruction because promptable 3D models often return plausible but unitless geometry. In the submitted \method workflow, visible plate diameter is the primary metric cue, and MoGe-2~\cite{moge22025} is used only as an auxiliary cue for initial dish-diameter estimation when direct plate measurement is uncertain. These cues support the central principle of \method: separate unitless shape generation from metric scale recovery. Table~\ref{tab:related_positioning} summarizes how this setting differs from static food-volume estimation, egocentric intake monitoring, and single-image 3D foundation-model reconstruction.

\begin{table}[t]
\centering
\caption{Positioning of \method relative to common food-volume and reconstruction settings. PerBite targets paired consumption states and explicitly separates unitless 3D shape from metric volume calibration.}
\label{tab:related_positioning}
\scriptsize
\setlength{\tabcolsep}{2.2pt}
\begin{tabularx}{\linewidth}{>{\raggedright\arraybackslash}Xcccc}
\toprule
\rowcolor{TableHeader}
\thead{Setting} & \thead{Input} & \thead{3D} & \thead{Metric} & \thead{$\Delta V$} \\
\modernrule
Static food-volume estimation & image/depth & partial & often & no \\
\rowcolor{TableStripe}
Egocentric intake monitoring & video & usually no & indirect & yes \\
Single-image 3D foundation models & image & yes & no & no \\
\rowcolor{TableStripe}
\method & before/after states & yes & yes & yes \\
\bottomrule
\end{tabularx}
\end{table}

\begin{figure*}[h]
\centering
\includegraphics[width=0.98\textwidth]{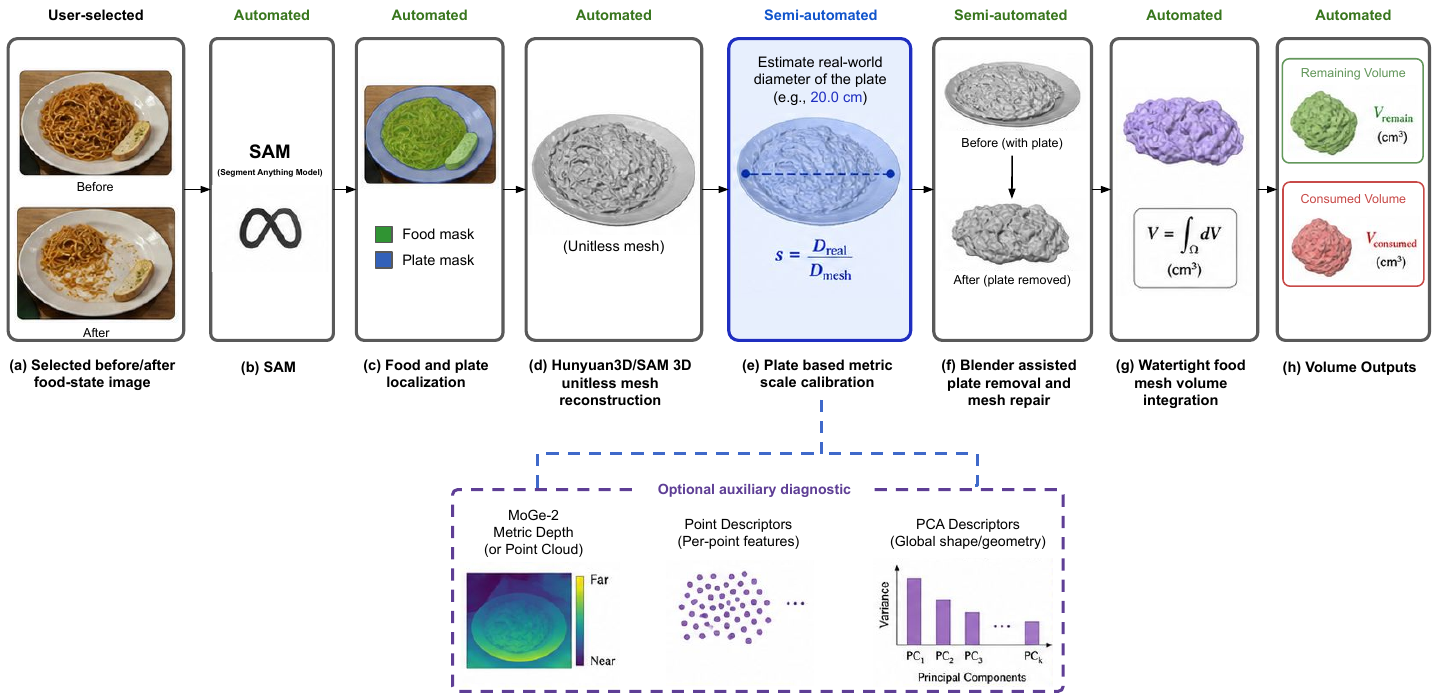}
\caption{Visualization of the PerBite pipeline for paired before- and after-consumption food-state volume estimation. (a) Selected before- and after-consumption food-state images used as pipeline inputs. (b) SAM 3 segmentation. (c) Food and plate localization for isolating the target food region prior to reconstruction. (d) Hunyuan3D/SAM3D-based unitless mesh reconstruction from the localized food crop. (e) Plate-based metric scale calibration for converting the reconstructed unitless mesh into physical dimensions. (f) Blender-assisted plate removal and mesh repair for refining the reconstructed geometry. (g) Watertight food-mesh volume integration for estimating the remaining food volume. (h) Final volume outputs in $cm^3$ for the before- and after-consumption states, which are subsequently used to estimate the volume of food consumed.
}
\label{fig:Perbite_pipeline}
\end{figure*}

\section{The Proposed Method}
\label{sec:method}

\begin{figure*}[t]
\centering
\resizebox{\textwidth}{!}{%

\begin{tikzpicture}[
    node distance=0.6cm and 0.4cm, 
    >=Latex, 
    every node/.style={font=\small\sffamily}
]

\tikzset{
    inputbox/.style={draw=inputtext, fill=inputbg, text=inputtext, rounded corners=3pt, align=center, font=\scriptsize\bfseries, minimum height=0.85cm, minimum width=1.65cm, line width=0.6pt},
    corebox/.style={draw=coretext, fill=corebg, text=coretext, rounded corners=3pt, align=center, font=\scriptsize\bfseries, minimum height=0.85cm, minimum width=1.65cm, line width=0.6pt},
    calibbox/.style={draw=calibtext, fill=calibbg, text=calibtext, rounded corners=3pt, align=center, font=\scriptsize\bfseries, minimum height=0.85cm, minimum width=1.65cm, line width=0.6pt},
    subbox/.style={draw=auxtext!30, fill=white, text=auxtext, rounded corners=3pt, align=center, font=\scriptsize, minimum height=0.75cm, minimum width=1.65cm, line width=0.5pt},
    groupcore/.style={draw=coretext!40, fill=corebg!12, dashed, rounded corners=4pt, line width=0.6pt, inner sep=0.15cm},
    groupcalib/.style={draw=calibtext!40, fill=calibbg!12, dashed, rounded corners=4pt, line width=0.6pt, inner sep=0.15cm},
    arrow/.style={->, line width=0.6pt, draw=black!50, shorten >=1pt, shorten <=1pt}
}

\node[inputbox] (state) {Selected state\\image};
\node[corebox, right=of state] (sam3) {SAM 3\\segmentation};
\node[subbox, right=of sam3] (mask) {food/plate\\masks \& crop};
\node[corebox, right=of mask] (sam3d) {Hunyuan3D/\\SAM 3D mesh};
\node[subbox, right=of sam3d] (cleanup) {Plate removal\\and repair};
\node[calibbox, right=of cleanup] (scale) {Metric scale\\plate\\diameter};
\node[calibbox, right=of scale] (vol) {Metric volume\\$\mathrm{Vol}(s\mathcal{M}^u)$};
\node[calibbox, right=of vol] (traj) {Bite-level\\trajectory};

\node[subbox, below=0.6cm of sam3] (prompt) {``food''/``plate''\\ prompt};
\node[subbox, below=0.6cm of scale] (moge) {Optional MoGe-2\\dish diameter};
\node[subbox, below=0.6cm of vol, xshift=0.95cm] (mono) {monotonic\\depletion check};

\draw[arrow] (state) -- (sam3);
\draw[arrow] (prompt) -- (sam3);

\draw[arrow] (sam3) -- (mask);
\draw[arrow] (mask) -- (sam3d);
\draw[arrow] (sam3d) -- (cleanup);
\draw[arrow] (cleanup) -- (scale);

\draw[arrow] (moge) -- (scale);
\draw[arrow] (scale) -- (vol);

\draw[arrow] (vol) -- (traj);
\draw[arrow] (vol) -- (mono);
\draw[arrow] (mono) -- (traj);

\begin{scope}[on background layer]
    \node[groupcore, fit=(sam3)(mask)(sam3d)(cleanup), label={[font=\scriptsize\bfseries, text=coretext]above:Promptable Reconstruction}] {};
    \node[groupcalib, fit=(scale)(moge), label={[font=\scriptsize\bfseries, text=calibtext]above:Metric Calibration}] {};
\end{scope}
\end{tikzpicture}%
}

\caption{Curated \method workflow used for the challenge submission. SAM~3 provides food and plate masks; Hunyuan3D/SAM~3D generates an initial unitless food mesh; the mesh is scaled using the plate diameter, with MoGe-2 used only as an auxiliary initial cue for dish diameter when needed; plate geometry is removed in Blender; and the remaining food mesh is hole-filled, made watertight, and integrated for volume. The same procedure is applied to the before/after states and evaluated using remaining-volume, consumed-volume, Chamfer, and monotonic-depletion diagnostics.}
\label{fig:pipeline}
\end{figure*}
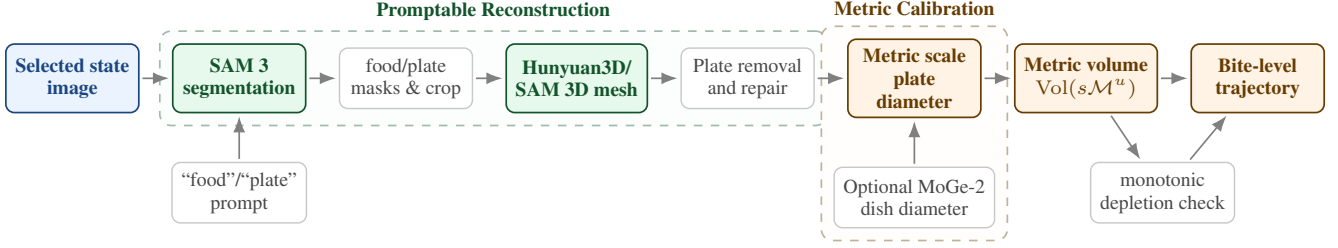

Our method starts from selected before- and after-consumption state images and follows five main stages: input state preparation, promptable food and plate localization with SAM~3~\cite{carion2025sam3}, Hunyuan3D/SAM~3D~\cite{zhao2025hunyuan3d,sam3d2025} unitless mesh reconstruction, metric scale calibration, and paired-state volume inference. Implementation-specific steps, including Blender-assisted plate/support removal, watertight mesh repair, and volume integration, are reported explicitly because they were part of the submitted challenge workflow and expose practical bottlenecks in current foundation-model reconstruction systems. Table~\ref{tab:notation} defines the main symbols used throughout the method, Figure~\ref{fig:Perbite_pipeline} visualizes the complete PerBite pipeline, Figure~\ref{fig:pipeline} summarizes the workflow, and Algorithm~\ref{alg:method} gives the inference procedure for one before/after pair.

\begin{table}[t]
\centering
\caption{Main symbols used by \method. Superscript $u$ denotes unitless mesh quantities, while metric volumes are reported in physical units.}
\label{tab:notation}
\scriptsize
\setlength{\tabcolsep}{3pt}
\begin{tabularx}{\linewidth}{lX}
\toprule
\rowcolor{TableHeader}
\thead{Symbol} & \thead{Meaning} \\
\midrule
$I_t$ & selected RGB image at food state $t$ \\
\rowcolor{TableStripe}
$M_{f,t}$, $M_{p,t}$ & food and plate masks at state $t$ \\
$\mathcal{M}^{u}_t$ & unitless mesh reconstructed from the selected state image \\
\rowcolor{TableStripe}
$s_t$ & uniform length-scale factor applied to the unitless mesh \\
$\mathcal{C}_t$ & cleaned food mesh after support removal and watertight repair \\
\rowcolor{TableStripe}
$\hat V_t$ & predicted remaining food volume at state $t$ \\
$\widehat{\Delta V}_t$ & predicted consumed volume between two states \\
\bottomrule
\end{tabularx}
\end{table}

\subsection{Problem Definition}
\label{sec:problem_def}
Given selected before/after food states from an eating sequence, \method estimates a sequence of remaining food volumes $\{\hat V_t\}_{t=0}^{T}$ and consumed increments $\widehat{\Delta V}_t=\hat V_{t-1}-\hat V_t$. For each evaluated food item, we form one before-consumption state image and one after-consumption state image from the egocentric sequence. This setup isolates the reconstruction, metric scaling, cleanup, and volume-integration problem; automatic temporal state selection is left as future work. The method should satisfy three requirements: (i) \emph{metric plausibility}, because generated meshes are unitless and must be converted to physical volume; (ii) \emph{food-state consistency}, because before/after states should describe the same physical item after consumption; and (iii) \emph{monotonic depletion consistency}, because remaining edible volume should not increase after a bite event except within uncertainty.

\subsection{Input State Images}
\label{sec:input_states}
\method operates on one RGB image per food state. The Kaggle challenge data are egocentric eating sequences rather than pre-cropped before-and-after-image pairs, so the two state images used for the reported submission were selected outside the released reconstruction pipeline: one frame representing the visible food before consumption and one frame after consumption, avoiding severe hand or utensil occlusion when possible. We do not claim automatic bite detection or automatic temporal frame selection as contributions in this paper, and the public code starts from the selected state images. This design keeps the evaluated pipeline focused on reconstruction, metric calibration, cleanup, volume integration, and paired-state diagnostics.

\subsection{Promptable Food and Plate Localization with SAM~3}
\label{sec:sam3_localization}
To make the segmentor more general, \method avoids the need for a separate detector or a food-specific segmentation model. It uses SAM~3~\cite{carion2025sam3} directly with text prompts for the food and plate regions. For the food region, we use
\begin{equation}
p_f = \text{``food''}.
\end{equation}
SAM~3 returns candidate masks for regions matching the text prompt; an analogous plate prompt is used to obtain the plate mask when the plate is visible. Our implementation adds only a lightweight convenience option, \texttt{mask\_mode}, on top of the official SAM~3 behavior: \texttt{single} keeps the highest-scoring returned mask, whereas \texttt{multi} saves all returned masks along with an additional merged mask. In the reported challenge submission, the food mask defines the crop passed to Hunyuan3D/SAM~3D, and the plate mask provides the reference region for metric scale and later plate cleanup. This design provides two practical advantages. First, it removes a detector-specific dependency, making the first stage a single promptable concept-segmentation call. Second, it avoids hard-coding food categories, which is important for unseen items such as mixed meals, pastries, and partially consumed leftovers.

The selected SAM~3 food mask $M_{f,t}$ defines a square crop with a small context margin around the food region, while the plate mask $M_{p,t}$ is used to estimate the plate diameter and to guide Blender-assisted plate removal when the generator reconstructs food and support geometry together. The crop policy is kept fixed across the before- and after-states to avoid introducing artificial scale changes. FoodMem and BenchSeg-style segmentors are useful comparison points for future analysis, but they are not required components of the submitted workflow~\cite{almughrabi2025foodmem,almughrabi2026benchseg}.

\subsection{Hunyuan3D/SAM~3D Unitless Mesh Reconstruction}
\label{sec:sam3d_mesh}
Given the selected food crop and mask, Hunyuan3D/SAM~3D reconstructs a textured object mesh
\begin{equation}
\mathcal{M}^{u}_t=(\mathcal{V}^{u}_t,\mathcal{F}^{u}_t),
\end{equation}
where the superscript $u$ emphasizes that the mesh is in arbitrary units. We use Hunyuan3D/SAM~3D as a shape reconstruction prior rather than as a metric measurement device. Although the generated mesh may provide a plausible food surface, its scale, thickness, underside structure, and object boundaries can be biased. It may also include parts of the supporting plate or small disconnected artifacts. Therefore, the generated mesh is treated only as an intermediate reconstruction: it is subsequently scaled, cleaned to remove non-food geometry, repaired, and integrated to obtain volume.

\subsection{Metric Scale Calibration}
\label{sec:metric_scale}
Because the reconstructed mesh is unitless, \method estimates a length-scale factor $s_t$, uniformly scales the generated mesh, and then measures the volume of the scaled watertight mesh, as shown in Eq.~\ref{eq:cubic}:
\begin{equation}
\hat V_t = \operatorname{Vol}(s_t\mathcal{M}^{u}_t).
\label{eq:cubic}
\end{equation}
For a uniform scale, this is equivalent to cubic scaling of the mesh volume; therefore, a small error in length scale is amplified by a cubic factor in volume. The theoretical consequence is simple but important: if the estimated length scale is biased by a factor $(1+\epsilon)$, the volume is biased approximately by $(1+\epsilon)^3$. Thus, a reconstruction that looks geometrically plausible can still be unusable for intake estimation if its metric scale is weakly constrained.

\paragraph{Plate-based scale.}
In the submitted challenge workflow, plate diameter is the primary metric scale cue. When a plate or another known-size cue is visible, Eq.~\ref{eq:plate} estimates the scale from the metric plate diameter $D^{m}*{plate,t}$ and its corresponding diameter $d^u*{plate,t}$ measured in the unitless reconstruction. MoGe-2 is used only as an auxiliary cue for initial dish-diameter estimation when direct plate measurement is uncertain and not as the primary scale source for the reported challenge results. The scale is
\begin{equation}
s^{plate}*t = \frac{D^{m}*{plate,t}}{d^{u}_{plate,t}}.
\label{eq:plate}
\end{equation}

\paragraph{Auxiliary MoGe-2/PCA scale diagnostic.}
This component was used only as an auxiliary diagnostic and for early dish-diameter estimation, not as the primary scale source for the reported challenge result. When explicit plate or reference segmentation is unreliable, \method can estimate scale by comparing robust object-size descriptors between two point sets: metric 3D points within the SAM~3 food mask and dimensionless vertices from the SAM~3D mesh. The metric point set is obtained by applying MoGe-2 to the RGB image and retaining valid 3D points whose pixels lie inside the object mask. Both point sets are filtered to keep only finite coordinates and summarized in a PCA-aligned coordinate system using four robust size descriptors: two percentile-based extents and two dimensions from a PCA-aligned inlier bounding box.

\begin{figure*}[t]
\centering
\setlength{\tabcolsep}{2pt}
\begin{tabular}{ccc}
\small Mask overlay & \small Metric depth & \small Used pixels \\
\includegraphics[width=0.32\textwidth]{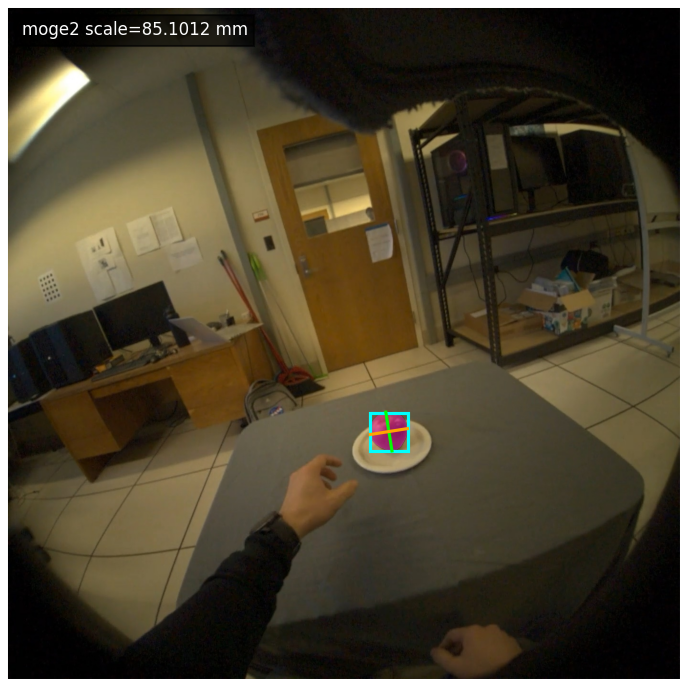} &
\includegraphics[width=0.32\textwidth]{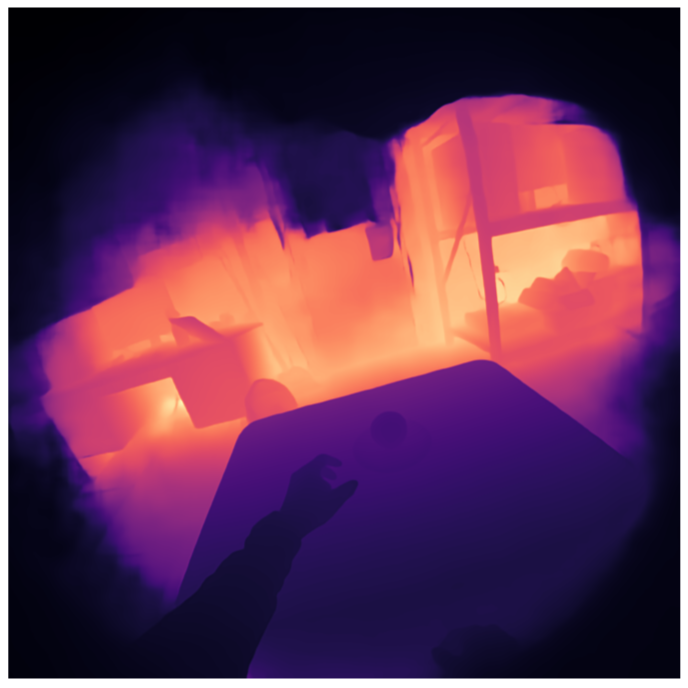} &
\includegraphics[width=0.32\textwidth]{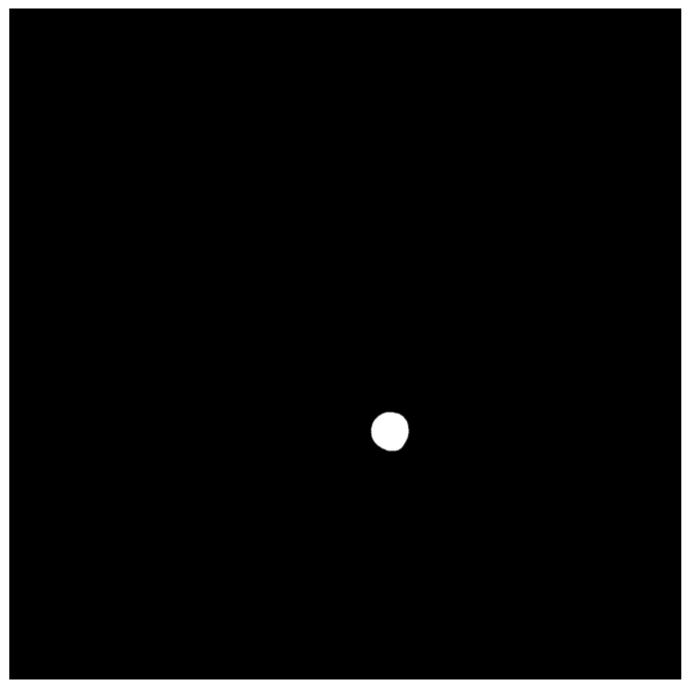} \\
\includegraphics[width=0.32\textwidth]{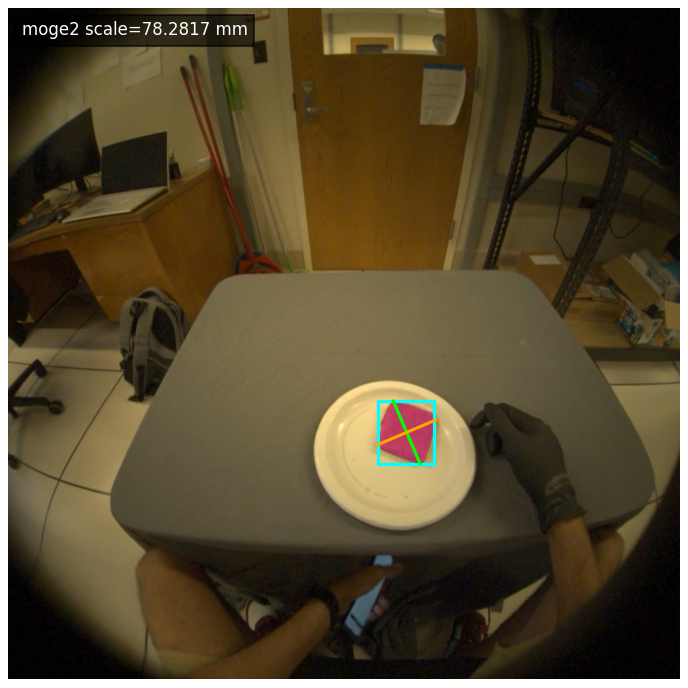} &
\includegraphics[width=0.32\textwidth]{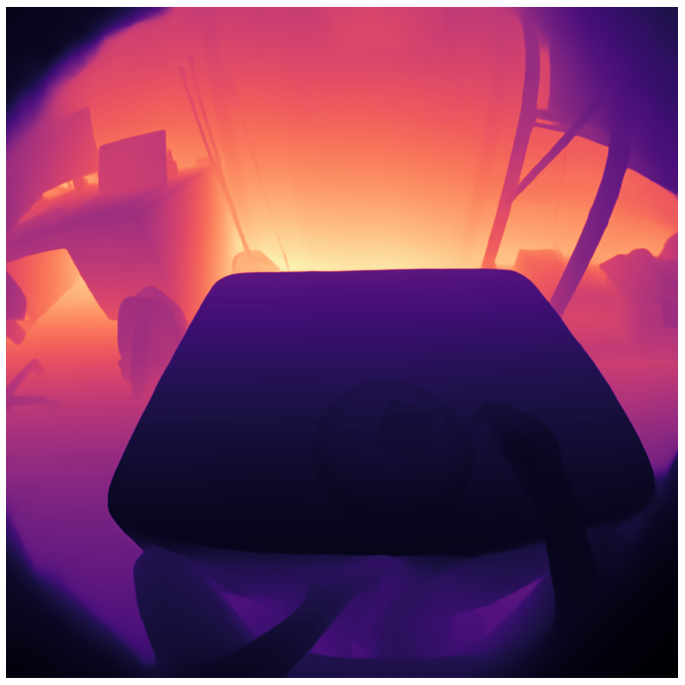} &
\includegraphics[width=0.32\textwidth]{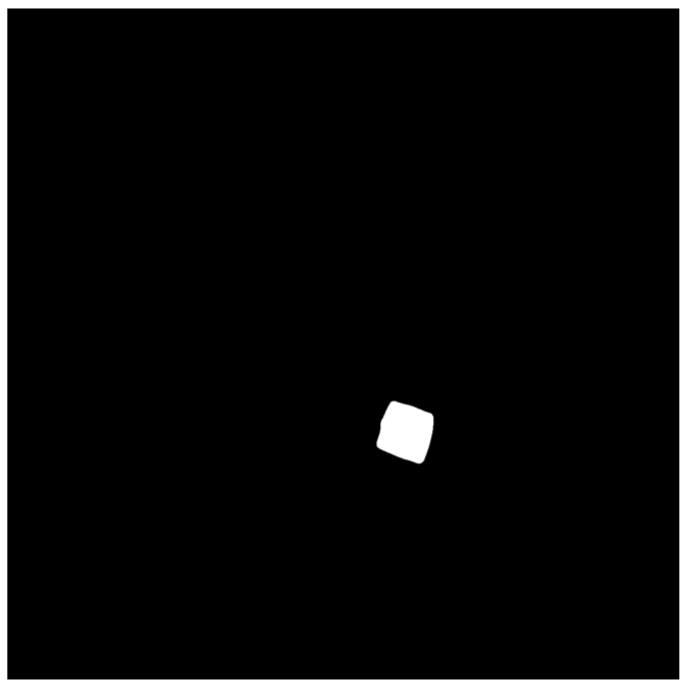} \\
\end{tabular}
\caption{Debug views for the auxiliary MoGe-2 to mesh scale diagnostic. Each row shows the SAM~3 mask overlay, MoGe-2 metric depth, and mask-valid pixels used for descriptor-based scale estimation. }
\label{fig:pca_scale_debug}
\end{figure*}

Let $D^m_{k,t}$ denote the descriptor $k$ computed from MoGe-2 metric object points after conversion to the desired output unit, and let $d^u_{k,t}$ denote the corresponding descriptor computed from the unitless mesh vertices. Each valid descriptor proposes a scale candidate
\begin{equation}
s_{k,t}=\frac{D^m_{k,t}}{d^u_{k,t}},
\label{eq:pca_candidate_scale}
\end{equation}
where candidates with non-finite or non-positive descriptors are discarded. Remaining candidates are filtered in log-space using a median absolute deviation rule, and the final scale is the median of the accepted candidates:
\begin{equation}
s^{pca}_t=\operatorname{median}_{k\in\mathcal{A}_t} s_{k,t},
\label{eq:pca_scale}
\end{equation}
where $\mathcal{A}_t$ is the accepted candidate set. The method applies a single uniform scale factor to the mesh; it neither deforms the mesh nor performs non-uniform axis-wise scaling. The descriptor family used by this optional diagnostic is summarized in Table~\ref{tab:pca_scale_descriptors}. Figure~\ref{fig:pca_scale_debug} visualizes representative SAM~3 masks, MoGe-2 depth, and valid pixels used by this estimator.

\begin{table}[t]
\centering
\caption{Descriptor-wise scale candidates used by the optional MoGe-2/PCA scale diagnostic. Superscript $m$ denotes metric MoGe-2 object points and superscript $u$ denotes unitless SAM~3D mesh vertices.}
\label{tab:pca_scale_descriptors}
\small

\begin{tabularx}{\linewidth}{Xcc}
\toprule
\rowcolor{TableHeader}
\thead{Descriptor} & \thead{Metric value} & \thead{Candidate scale} \\
\modernrule
PCA major-axis extent & $D^{m}_{\mathrm{maj}}$ & $D^{m}_{\mathrm{maj}}/d^{u}_{\mathrm{maj}}$ \\
\rowcolor{TableStripe}
PCA mid-axis extent & $D^{m}_{\mathrm{mid}}$ & $D^{m}_{\mathrm{mid}}/d^{u}_{\mathrm{mid}}$ \\
PCA-box long dimension & $B^{m}_{\mathrm{long}}$ & $B^{m}_{\mathrm{long}}/b^{u}_{\mathrm{long}}$ \\
\rowcolor{TableStripe}
PCA-box mid dimension & $B^{m}_{\mathrm{mid}}$ & $B^{m}_{\mathrm{mid}}/b^{u}_{\mathrm{mid}}$ \\
\bottomrule
\end{tabularx}
\end{table}

\paragraph{Scale confidence and cue selection.}
When a reliable plate/reference cue is available, we use it as the primary scale cue; this is the configuration corresponding to the reported challenge results. The MoGe-2/PCA descriptor scale is retained as an auxiliary diagnostic. If both cues are available, we compute both and use their disagreement as a diagnostic rather than silently averaging incompatible measurements. Specifically, we define
\begin{equation}
u_s(t)=\left|\log s^{plate}_t-\log s^{pca}_t\right|,
\end{equation}
which flags unreliable predictions when the two metric cues disagree. For the PCA descriptor cue itself, we report the number of accepted descriptor candidates and the relative interquartile range $\mathrm{IQR}(\{s_{k,t}\}_{k\in\mathcal{A}_t})/s^{pca}_t$ as a rule-based confidence diagnostic. Low relative spread indicates that independent major-axis, mid-axis, and PCA-aligned OBB descriptors agree on the metric scale. We do not assume that MoGe-2 provides perfect object geometry; it is used only as a metric size cue over mask-valid pixels, and candidate-scale spread exposes when this cue is unreliable.

\subsection{Temporal Consistency and Repeated Inference}
\label{sec:temporal_consistency}
SAM~3D reconstruction can be stochastic. When multiple stochastic reconstructions are available, the implementation can average the scaled volumes:
\begin{equation}
\bar V_t = \frac{1}{R}\sum_{r=1}^{R} \operatorname{Vol}(s_{t,r}\mathcal{M}^{u}_{t,r}).
\end{equation}
For a before/after pair, \method predicts consumed volume as
\begin{equation}
\widehat{\Delta V}_t=\max(0,\bar V_{t-1}-\bar V_t).
\end{equation}
Monotonicity is evaluated on the raw before/after volume estimates before this non-negative clipping. For longer sequences, we optionally estimate a physically consistent trajectory using the monotonic-depletion objective in Eq.~\ref{eq:mono}:
\begin{equation}
\min_{V_0,\ldots,V_T} \sum_{t=0}^{T} \rho(V_t-\bar V_t) + \gamma \sum_{t=1}^{T}\max(0, V_t-V_{t-1})^2,
\label{eq:mono}
\end{equation}
where $\rho$ is a robust loss, and the second term penalizes non-monotone increases in volume. In the reported paired-state results, all 17 before/after predictions already satisfy monotonic depletion, so no correction is needed to obtain the reported violation rate.

\subsection{Implementation Details and Reproducibility}
\label{sec:implementation_details}
Unless stated otherwise, the reported configuration uses SAM~3 food and plate masks, a fixed crop policy for the before/after states of each item, Hunyuan3D/SAM~3D mesh generation, metric scaling from plate diameter, Blender-assisted removal of plate/support geometry, hole filling, watertight repair, and signed mesh-volume computation. MoGe-2 is retained only as an auxiliary cue for initial dish-diameter estimation and as a scale-disagreement diagnostic; it is not the primary scale source for the headline challenge numbers. When repeated stochastic outputs are generated, the implementation averages the scaled volumes; otherwise, it reports the single scaled reconstruction. For reproducibility, the implementation records the SAM~3 and Hunyuan3D/SAM~3D checkpoints, prompts, crop margin, image resolution, Blender cleanup steps, watertight post-processing settings, mesh-volume operator, and plate-diameter scale used for each state.

\begin{algorithm}[t]
\caption{\method inference for one before/after food pair}
\label{alg:method}
\begin{algorithmic}[1]
\STATE Input: before and after images $\{I_{before},I_{after}\}$, food prompt $p_f=\text{``food''}$, plate prompt when visible
\FOR{each state $t\in\{\mathrm{before},\mathrm{after}\}$}
    \STATE Run SAM~3 to obtain food mask $M_{f,t}$ and plate mask $M_{p,t}$ when visible
    \STATE Crop the food region and reconstruct unitless mesh $\mathcal{M}^{u}_t$ with Hunyuan3D/SAM~3D
    \STATE Estimate length scale $s_t$ from plate diameter; optionally use MoGe-2 only to initialize or check dish diameter
    \STATE Uniformly scale the mesh, remove plate/support geometry in Blender, repair holes, and make the food mesh watertight to obtain $\mathcal{C}_t$
    \STATE Compute remaining volume $\hat V_t=\operatorname{Vol}(s_t\mathcal{C}_t)$
\ENDFOR
\STATE Compute raw consumed volume $\widehat{\Delta V}_{raw}=\hat V_{before}-\hat V_{after}$
\STATE Optionally report non-negative intake $\widehat{\Delta V}=\max(0,\widehat{\Delta V}_{raw})$
\STATE Output remaining volumes, consumed volume, Chamfer, MAPE, and monotonicity diagnostics
\end{algorithmic}
\end{algorithm}

\section{Experiments}
\label{sec:experiments}

This section reports the results for the completed challenge and diagnostic volume. We use the available metrics to analyze surface quality, metric volume accuracy, consumed-volume error, and physical consistency.

\subsection{Dataset and Evaluation Setting}
\label{sec:dataset_eval}
\begin{figure*}[t]
\centering
\includegraphics[width=0.98\textwidth]{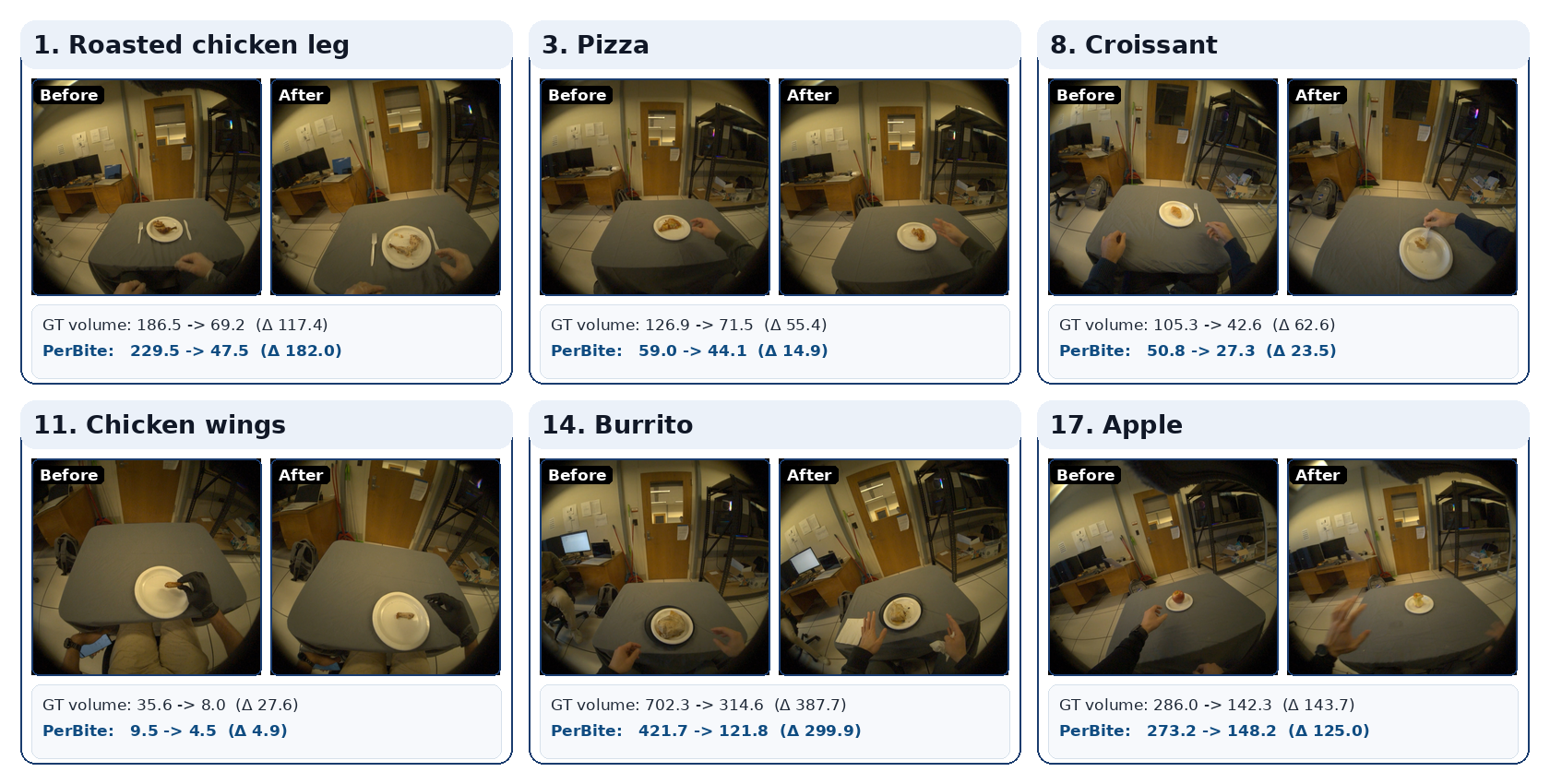}
\caption{Representative before/after food-state pairs with ground-truth and \method-predicted volumes. Each panel reports before $\rightarrow$ after volume and consumed-volume difference $\Delta$ in challenge volume units. The examples illustrate why \method evaluates selected paired states rather than independent static food images: viewpoint and hand position can vary while the edible item remains the same physical object.}
\label{fig:before_after_examples}
\end{figure*}

We evaluate \method on 17 before-and-after food pairs from the continuous 3D reconstruction-while-eating setting. Because this is the official hidden/test evaluation size, we treat the results as challenge validation and diagnostic evidence rather than broad statistical generalization. Each pair comprises a pre-consumption state, a post-consumption state, a predicted mesh for each state, and ground-truth before- and after-volumes. The evaluated foods include roasted chicken leg, quesadilla, pizza, muffin, hotdog, grilled salmon, garlic bread, croissant, crab cake, cinnamon roll, chicken wings, chicken sandwich, cheesecake, burrito, breaded fish, bagel, and apple. This paired design directly tests the core question of bite-aware volume estimation: can a method estimate both remaining volume at each state and consumed volume between states? Figure~\ref{fig:before_after_examples} illustrates representative paired states together with ground-truth and predicted volume changes.

The 2026 Kaggle-hosted MetaFood challenge setting differs from earlier static MTF/MetaFood reconstruction benchmarks. It focuses on continuous 3D reconstruction during eating: methods operate on egocentric eating observations and produce reconstructed meshes of food states before and after consumption. The before/after difference is intended to reflect food consumed during the eating episode. Unlike earlier controlled settings with explicit scale references, this setting stresses implicit metric reasoning from the image content, the reconstructed mesh, monocular geometry, and common food/container cues. Our evaluation, therefore, reports both the official surface-reconstruction score and the downstream volume diagnostics.

\paragraph{Official challenge protocol.}
The MetaFood CVPR 2026 Continuous 3D Reconstruction While Eating Challenge evaluates reconstructed food meshes against reference geometry for paired before- and after-consumption states. The official score is the average Chamfer distance after rigid ICP alignment. Importantly, ICP is restricted to rotation and translation; no scale correction is applied. This makes the metric sensitive to metric-size errors, which is appropriate for food-volume estimation because scale is part of the task rather than a nuisance variable.

Unless noted otherwise, the reported headline numbers correspond to the submitted curated \method configuration: SAM~3 food/plate segmentation, Hunyuan3D/SAM~3D mesh generation, plate-diameter scaling, Blender plate removal, watertight repair, and volume computation.

Figure~\ref{fig:before_after_examples} shows representative before/after states from the evaluated set. These images strengthen the experimental motivation: the task is not only object reconstruction but also state-change reconstruction under egocentric viewpoint variation, including hands, utensils, and partially consumed food geometry.

\begin{table}[t]
\centering
\caption{Official MetaFood CVPR 2026 challenge outcome. \method ranked first under the official protocol. The score is computed after rigid ICP alignment without scale correction.}
\label{tab:challenge_result}
\small
\begin{tabularx}{\linewidth}{>{\raggedright\arraybackslash}Xccc}
\toprule
\rowcolor{TableHeader}
\thead{Method} & \thead{Rank} & \thead{Official score\metricdown} & \thead{Alignment} \\
\modernrule
\method & \textbf{1st} & \textbf{8.31} & Rigid ICP, no scale \\
\bottomrule
\end{tabularx}
\end{table}
Table~\ref{tab:challenge_result} reports the official challenge outcome. The full leaderboard entries were not released for redistribution; therefore, we avoid reconstructing unofficial comparisons and report only the official rank and score of our submission. Table~\ref{tab:metric_summary} then separates the official surface-reconstruction score from the dietary-volume diagnostics used in this paper.

\begin{table}[t]
\centering
\caption{Compact view of the main evaluation signals. The challenge score measures surface reconstruction under the official no-scale ICP protocol, while the volume metrics evaluate dietary relevance.}
\label{tab:metric_summary}
\small
\begin{tabularx}{\linewidth}{>{\raggedright\arraybackslash}Xcc}
\toprule
\rowcolor{TableHeader}
\thead{Evaluation target} & \thead{Metric} & \thead{Value} \\
\modernrule
Challenge geometry & CD, rigid ICP/no scale\metricdown & \textbf{8.31} \\
\rowcolor{TableStripe}
Remaining volume & State MAPE\metricdown & 33.87\% \\
Consumed volume & $\Delta V$ MAPE\metricdown & 53.74\% \\
\rowcolor{TableStripe}
Physical trajectory & MVR\metricdown & \textbf{0.0\%} \\
\bottomrule
\end{tabularx}
\end{table}

\subsection{Metrics}
\label{sec:metrics}
We report four families of metrics. Because percentage error can be unstable when the true consumed volume is small, we report MAE and RMSE alongside MAPE and median APE rather than relying on a single percentage metric.

\paragraph{Surface reconstruction.}
For each reconstructed mesh, we report the Chamfer distance (CD) to the reference geometry, with lower values indicating better agreement. For challenge scoring, predicted and reference meshes are aligned using \emph{rigid} ICP only: rotation and translation are estimated, but no scale correction is applied. This makes the metric sensitive to metric-size errors, which is appropriate for volume reconstruction because scale is part of the task rather than a nuisance variable.

\paragraph{State-level volume.}
For before and after states, we report mean absolute percentage error (MAPE), median APE, mean absolute error (MAE), and root mean squared error (RMSE):
\begin{equation}
\mathrm{APE}_t = \frac{|\hat V_t - V_t|}{V_t+\epsilon}\times 100.
\end{equation}

\paragraph{Consumed volume.}
For each before/after pair, the consumed volume is
\begin{equation}
\Delta V = V_{before}-V_{after},\quad
\widehat{\Delta V}=\hat V_{before}-\hat V_{after}.
\end{equation}
We report the error on $\Delta V$ separately because it compounds errors from two reconstructed states.

\paragraph{Monotonic depletion.}
A monotonicity violation occurs if $\hat V_{after}>\hat V_{before}$. We report the violation count and rate:
\begin{equation}
\mathrm{MVR}=\frac{1}{N}\sum_{i=1}^{N}\mathbb{1}[\hat V^{(i)}_{after}>\hat V^{(i)}_{before}].
\end{equation}

\subsection{Chamfer Distance Results}
\label{sec:chamfer_results}
Table~\ref{tab:chamfer_results} reports surface reconstruction quality over 34 reconstructed meshes, while Table~\ref{tab:chamfer_extremes} highlights the best and worst object-state cases. \method obtains an average Chamfer distance of 8.31 when ICP scale correction is disabled. This score is higher than a scale-normalized alignment would produce, but it is the appropriate evaluation for metric food-volume estimation because scale is part of the problem rather than a nuisance variable. After-consumption states are slightly easier on average than before states, possibly because partially consumed objects are smaller or have less complex visible geometry. However, object-specific failures remain, especially for soft, elongated, or highly irregular foods.

\begin{table}[t]
\centering
\caption{Chamfer distance of \method reconstructions across before/after food states. CD is computed after rigid ICP alignment without scale correction. Lower is better.}
\label{tab:chamfer_results}
\small
\begin{tabularx}{\linewidth}{>{\raggedright\arraybackslash}Xccc}
\toprule
\rowcolor{TableHeader}
\thead{Split} & \thead{Meshes} & \thead{Mean CD\metricdown} & \thead{Median CD\metricdown} \\
\modernrule
Before states & 17 & 8.87 & 8.52 \\
\rowcolor{TableStripe}
After states & 17 & \textbf{7.76} & \textbf{6.42} \\
All states & 34 & 8.31 & 7.27 \\
\bottomrule
\end{tabularx}
\end{table}

\begin{table}[t]
\centering
\caption{Best and worst Chamfer cases under rigid ICP without scale correction. Worst cases typically involve soft, elongated, or boundary-ambiguous geometry.}
\label{tab:chamfer_extremes}
\small
\begin{tabularx}{\linewidth}{>{\raggedright\arraybackslash}Xc}
\toprule
\rowcolor{TableHeader}
\thead{Object-state} & \thead{CD\metricdown} \\
\modernrule
Chicken Wings after & \textbf{3.59} \\
\rowcolor{TableStripe}
Apple before & 3.63 \\
Grilled Salmon after & 3.83 \\
\rowcolor{TableStripe}
Muffin before & 4.32 \\
Grilled Salmon before & 4.47 \\
\midrule
\rowcolor{TableStripe}
Burrito after & 17.56 \\
Bagel before & 16.38 \\
\rowcolor{TableStripe}
Burrito before & 14.45 \\
Cheesecake after & 13.43 \\
\rowcolor{TableStripe}
Cheesecake before & 13.11 \\
\bottomrule
\end{tabularx}
\end{table}

\subsection{Volume Estimation Results}
\label{sec:volume_results}
Table~\ref{tab:volume_results} reports volume accuracy. Across all before-and-after states, \method achieves 33.87\% MAPE and 35.52\% median APE. Before states have slightly lower MAPE than after states, while after states have lower absolute error because their volumes are smaller.

\begin{table}[t]
\centering
\caption{Volume estimation on 17 before/after food pairs. Consumed-volume error is higher because $\Delta V$ compounds two state estimates.}
\label{tab:volume_results}
\scriptsize
\setlength{\tabcolsep}{2.2pt}
\renewcommand{\arraystretch}{1.08}
\begin{tabularx}{\linewidth}{>{\raggedright\arraybackslash}Xcccc}
\toprule
\rowcolor{TableHeader}
\thead{Eval.} &
\thead{MAPE\metricdown} &
\thead{Med. APE\metricdown} &
\thead{MAE\metricdown} &
\thead{RMSE\metricdown} \\
\modernrule
All states & 33.87 & 35.52 & 48.25 & 77.00 \\
\rowcolor{TableStripe}
Before only & \textbf{31.97} & \textbf{35.00} & 63.49 & 94.52 \\
After only & 35.78 & 36.04 & \textbf{33.01} & \textbf{54.07} \\
\rowcolor{TableStripe}
Consumed $\Delta V$ & 53.74 & 56.56 & 46.52 & 65.19 \\
\bottomrule
\end{tabularx}
\vspace{-0.5em}
\end{table}

The gap between state-level volume error and consumed-volume error is the most important empirical finding. Although \method produces plausible state-level volume estimates, the bite amount $\Delta V$ is more sensitive because errors in $\hat V_{before}$ and $\hat V_{after}$ can add rather than cancel. The absolute error remains too high for deployment as an automatic dietary-monitoring system; the value of this result is diagnostic. This finding reinforces the argument that surface quality alone is not enough for dietary volume estimation; scale recovery, controlled mesh cleanup, watertightness, and volume-difference stability must be evaluated explicitly.

\subsection{Monotonic Depletion Results}
\label{sec:monotonicity_results}
\begin{figure}[t]
\centering
\includegraphics[width=\linewidth]{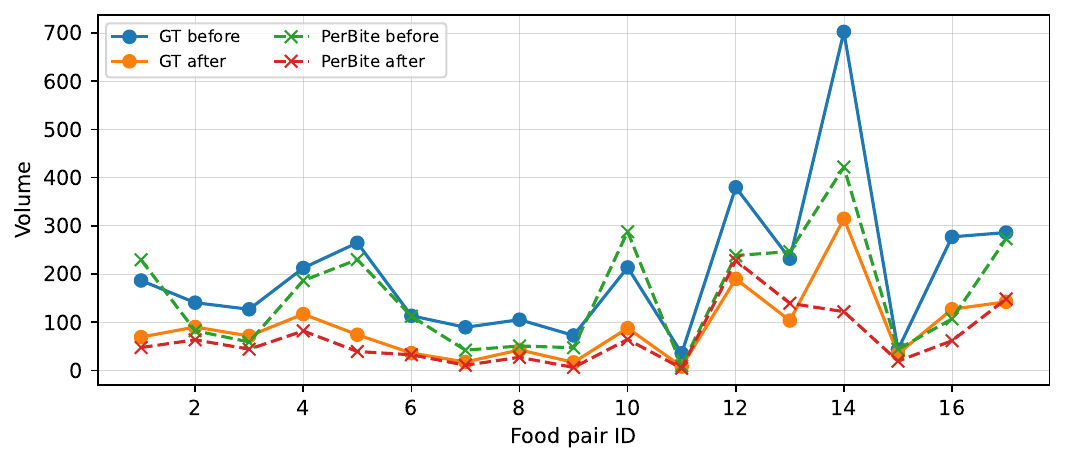}
\caption{Ground-truth and predicted before/after volumes for 17 food pairs. \method preserves the before-to-after decrease for every pair, but several items show large-scale or shape-induced volume error.}
\label{fig:volume_trajectory}
\end{figure}

Table~\ref{tab:monotonicity_results} reports monotonic depletion consistency. \method produces no violations across the 17 before/after pairs: every predicted after-volume is lower than its corresponding predicted before-volume. This is important as a necessary physical-consistency check, but it should not be interpreted as sufficient evidence of accurate intake estimation, because the magnitude of consumed volume remains imperfect. Figure~\ref{fig:volume_trajectory} visualizes the predicted and ground-truth before/after volumes for all 17 pairs.

\begin{table}[t]
\centering
\caption{Bite-wise monotonic depletion consistency. A violation occurs when the predicted after-consumption volume exceeds the predicted before-consumption volume.}
\label{tab:monotonicity_results}
\small
\begin{tabularx}{\linewidth}{Xcc}
\toprule
\rowcolor{TableHeader}
\thead{Pairs} & \thead{Violations\metricdown} & \thead{Violation rate\metricdown} \\
\modernrule
17 & \textbf{0} & \textbf{0.0\%} \\
\bottomrule
\end{tabularx}
\end{table}

\subsection{Relationship Between Surface and Volume Accuracy}
\label{sec:surface_volume_relation}
Chamfer distance and volume error do not necessarily correlate. In our results, the empirical correlation between per-state Chamfer distance and volume APE is approximately $0.41$; this value should be interpreted with caution because the official evaluation includes only 34 state meshes. This moderate relationship confirms that poor surface reconstruction often reduces volume, but it also shows that CD alone is insufficient: a mesh can be close in surface distance yet still produce poor metric volume if its scale, thickness, watertightness, or unobserved geometry is incorrect. This observation motivates reporting both surface and volumetric evaluation.

\section{Discussion and Limitations}
\label{sec:discussion}
\noindent
This section summarizes what the challenge result validates and where \method remains limited, especially for metric scale, surface-to-volume consistency, and consumed-volume estimation.

\subsection{Challenge Validation and Main Contribution}
\label{sec:discussion_challenge}
The first-place challenge result shows that a compact SAM~3 plus Hunyuan3D/SAM~3D workflow, when paired with plate-based metric calibration and careful mesh cleanup, can be competitive under a fixed surface-reconstruction protocol. The main contribution, however, is not only the rank. \method combines unitless reconstruction, plate-diameter scaling, Blender-assisted plate removal, watertight volume integration, paired-state volume estimation, and monotonic-depletion checks into a diagnostic framework for bite-aware intake. This framing makes the workflow easier to audit: failures can be assigned to prompt localization, shape completion, plate-scale calibration, controlled mesh cleanup, watertightness, mask/crop quality, or volume differencing.

\subsection{Why Surface Accuracy Is Not Enough}
\label{sec:discussion_surface_volume}
The chamfer distance and volume error diverge because they assess different properties. A mesh may be close in surface distance but still exhibit incorrect global scale, thickness, watertightness, or a hidden underside. The observed CD--volume correlation of approximately $0.41$ therefore supports the central message: dietary reconstruction should report surface distance and metric volume, not one as a proxy for the other.

\subsection{Why Consumed Volume Is Harder}
\label{sec:discussion_consumed}
State-level volume reaches 33.87\% MAPE, while consumed-volume error is 53.74\%. This gap is expected because the consumed volume is the difference between the two state estimates. Even moderate before/after errors can accumulate if their signs differ. Future improvements should therefore directly optimize paired-state scale consistency, rather than treating the before- and after-reconstructions as independent predictions.

\subsection{Scale Evidence and Failure Modes}
\label{sec:discussion_scaling}
Plate diameter is the primary metric cue in the submitted workflow, making the final volume sensitive to plate segmentation, diameter estimation, and the correspondence between the image-space plate and the generated mesh scale. MoGe-2/PCA descriptors are useful only as auxiliary dish-diameter checks or diagnostics when the plate cue is uncertain, but they can fail when visible metric points cover only part of the food while the generator hallucinates hidden geometry. Burrito, Bagel, and Cheesecake are the most difficult cases: elongated, ring-like, soft, or boundary-ambiguous foods amplify small errors in thickness, closure, cleanup, and scale.

\subsection{Limitations}
\label{sec:discussion_limits}
The current system does not automatically select before/after frames from raw video. The released pipeline starts from selected state images and focuses on promptable reconstruction, plate-based scale recovery, controlled mesh cleanup, watertight repair, and volume diagnostics. The current evaluation uses 17 paired before/after states rather than full long-horizon multi-bite videos; thus, the results validate paired state differencing and diagnostic volume evaluation, but not full long-horizon temporal tracking. The current workflow should be interpreted as a curated reconstruction protocol rather than a deployed fully automatic dietary-monitoring system: plate/support geometry is removed with Blender assistance, and hole filling and watertight repair are required before volume computation. The method can also fail when SAM~3 merges food with the container, when Hunyuan3D/SAM~3D changes object thickness, when the plate diameter is uncertain, or when the food has hidden cavities or a strong nonconvex structure. Bowls, soups, reflective plates, and heavily occluded foods require separate analysis.

\section{Conclusions}
\label{sec:conclusion}
\method demonstrates that promptable single-image 3D reconstruction can support bite-aware food-volume diagnostics when unitless shape generation is decoupled from plate-based metric-scale recovery and is followed by explicit mesh cleanup. The curated workflow ranks first under the official challenge surface metric, produces monotonic before-and-after volume predictions across the 17 evaluated pairs, and reveals the harder problem of estimating consumed volume. Rigid ICP without scale correction is stricter than scale-normalized alignment because metric scale is part of the food-volume task, not a nuisance variable. The main lesson is that food reconstruction benchmarks should report metric scale, cleanup assumptions, watertightness, and consumed-volume diagnostics alongside surface distance, because these quantities can diverge substantially in dietary assessment.

\section{Acknowledgments}
This work has been partially supported by the PID2022-141566NBI00 (AEI-MICINN), AIA2025-163919-C51 funded by Agencia Estatal de Investigación (AEI), Ministerio de Ciencia, Innovación y Universidades, Spain, and Icrea Academia’2022 (Generalitat de Catalunya). It has also been partially funded by CNS2022-135480 (A-BMC) funded by MICIU/AEI/ 10.13039/501100 011033, by FEDER (UE), and by European Union NextGenerationEU/ PRTR. A. AlMughrabi acknowledges the support of FPI Becas, MICINN, Spain. U. Haroon acknowledges the support of FI-SDUR Becas, MICINN, Spain. F. Al-Areqi acknowledges the support of the FI 2025 doctoral fellowship (AGAUR), Spain. The authors thankfully acknowledge the RES resources provided by the Barcelona Supercomputing Center on MareNostrum5 for IM-2025-3-0008.

{
    \small
    \bibliographystyle{ieeenat_fullname}
    \bibliography{main}
}
\end{document}